# Performance Analysis of Machine Learning Techniques to Predict Diabetes Mellitus

Md. Faisal Faruque
*Department of Computer Science & Engineering*
*Chittagong University of Engineering & Technology*
Chittagong, Bangladesh
faisal_uits@yahoo.com

Asaduzzaman
*Department of Computer Science & Engineering*
*Chittagong University of Engineering & Technology*
Chittagong, Bangladesh
asad@cuet.ac.bd

Iqbal H. Sarker
*Department of Computer Science & Engineering*
*Chittagong University of Engineering & Technology*
Chittagong, Bangladesh
iqbal@cuet.ac.bd

*Abstract*—Diabetes mellitus is a common disease of human body caused by a group of metabolic disorders where the sugar levels over a prolonged period is very high. It affects different organs of the human body which thus harm a large number of the body's system, in particular the blood veins and nerves. *Early prediction* in such disease can be controlled and save human life. To achieve the goal, this research work mainly explores various *risk factors* related to this disease using machine learning techniques. Machine learning techniques provide efficient result to extract knowledge by constructing predicting models from diagnostic medical datasets collected from the diabetic patients. Extracting knowledge from such data can be useful to predict *diabetic* patients. In this work, we employ four popular *machine learning algorithms*, namely Support Vector Machine (SVM), Naive Bayes (NB), K-Nearest Neighbor (KNN) and C4.5 Decision Tree, on adult population data to predict diabetic mellitus. Our experimental results show that C4.5 decision tree achieved higher accuracy compared to other machine learning techniques.

## I. INTRODUCTION

Diabetes mellitus, also known as diabetic, is a disease that affects the hormone insulin, resulting in abnormal metabolism of carbohydrates and improve levels of sugar in the blood. This high blood sugar affects various organs of the human body which in turn complicates many cause of the body, in particular the blood veins and nerves. The causes of diabetic is not yet completely discovered, many researchers believed that both hereditary elements and environmental factors are involved therein. In any case, diabetic can used to be most common in grown-ups and that's why it called "adult-onset" diabetes. It is currently trusted that diabetes mellitus is especially involved with the ageing process.

As indicated by Canadian Diabetes Association (CDA), somewhere in the year of 2010 to 2020, the quantity of individual person figure out to have diabetic in Canada is relied upon to escalate from 2.5 million to around 3.7 million [6]. The current worldwide situation is not different from this. As indicated by the International Diabetes Federation, number of people having diabetes mellitus achieved 382 million out of 2013 [7] that bring 6.6% of the world's total grown-up population. According to the world healthcare medical data it has been expected that diabetic disease will be increase from 376 billion to 490 billion within the year 2030 [8]. Moreover, diabetic is a conceivably independent contributing risk factor to micro-vascular entanglements. Diabetic patients are probably more powerless against a hoisted risk of micro-vascular damage, in this way long term complication affects of cardio-vascular disease is the leading *cause of death*. This micro-vascular harm and hasty cardio vascular disease eventually prompt to retinopathy, nephropathy and neuropathy [9].

Early prediction of such disease can be controlled over the diseases and save human life. To achieve this goal, this research work mainly explores the *early prediction of diabetes* by taking into account various *risk factors* related to this disease. For the purpose of the study we collect diagnostic dataset having 16 attributes diabetic of 200 patients. These attributes are age, diet, hyper-tension, problem in vision, genetic etc. In later part, we discuss about these attributes with their corresponding values. Based on these attributes, we build prediction model using various *machine learning* techniques to predict diabetes mellitus.

Machine learning techniques provide efficient result to extract knowledge by constructing predicting models from diagnostic medical datasets collected from the diabetic patients. Extracting knowledge from such data can be useful to predict diabetic patients. Various machine learning techniques have the ability to predict diabetes mellitus. However it is very difficult to *choose* the best technique to predict based on such attributes. Thus for the purpose of the study, we employ four popular machine learning algorithms, namely Support Vector Machine (SVM), Naive Bayes (NB), K-Nearest Neighbor (KNN) and C4.5 decision tree, on adult population data to predict diabetic mellitus.

The contributions in this study are as below-

- We *collect* real diagnostic dataset having various attributes or risk factors of diabetes mellitus of 200 patients.
- We make *performance comparison* of different machine learning techniques and evaluate the prediction results based on the relevant risk factors.

The rest part of this paper is composed as follows. Section II reviews the related works. We present our methodology in Section III. In Section IV, we report our experimental results. Finally, we conclude this paper in Section V.

## II. RELATED WORK

Many researchers have been conducted studies in the area of diabetic by using machine learning techniques to extract knowledge from available medical data. For instance, ALjumah et. al. [10] developed a predictive analysis model using support vector machine algorithm. In [11], Kavakiotis et al. used 10 fold cross validation as a evaluation method in three different algorithms, including Logistic regression, Naive Bayes, and SVM, where SVM provided better performance and accuracy of 84 % than other algorithms. In [12], Zheng et al. applied Random Forest, KNN, SVM, Naive Bayes, decision tree and logistic regression to predict diabetes mellitus at early stage, where filtering criteria can be improved. Swarupa et al. [13] applied KNN, J48, ANN, ZeroR and NB on various diabetes dataset. Pradeep et al. [14] applied KNN, J48, SVM and Random Forest, where J48 machine learning algorithm provides better performance and accuracy than others before preprocessing technique. The classification algorithms did not evaluate using cross validation method. To predict and control diabetes mellitus Huang et al [15] discussed three data mining techniques, including IB1, Naïve Bayes and C4.5 on dataset gathered from Ulster Community and Hospitals Trust (UCHT) in the year of 2000 to 2004. By applying feature selection technique, the performance of IB1 and Naive Bayes provided better result. In [16], Xue-Hui Meng et al. used three different data mining techniques ANN, Logistic regression, and J48 to predict the diabetic diseases using real world data sets by collecting information by distributed questioner. Finally it was concluded as J48 machine learning techniques provided efficient and better accuracy than others.

In this work, we analyze real diagnostic medical data based on various *risk factors* using machine learning classification techniques to evaluate their performance for predicting diabetes mellitus.

III. METHODOLOGY

In order to achieve our goal, study methodology comprises of few stages, which are accumulation of diabetes dataset with the relevant attributes of the patients, preprocessing the numeric value attributes, to apply various machine learning classification techniques and corresponding predictive analysis utilizing such data. In the following, we briefly discuss these phases.

*A. Dataset and Attributes*

In this work, we collect diabetes data from the diagnostic of Medical Centre Chittagong (MCC), Bangladesh. The dataset consists of various attributes or *risk factors* of diabetes mellitus of 200 patients. We have summarized the attributes and corresponding values in Table 1.

**Table 1: Dataset Description**

| SI No. | Attributes | Type | Values |
|---|---|---|---|
| 1 | Age (Years) | Numeric | {1 to 100} |
| 2 | Sex | Nominal | {Male, Female} |
| 3 | Weight (Kg's) | Numeric | {5 to 120} |
| 4 | Diet | Nominal | {Vegetarian, Non-Vegetarian} |
| 5 | Polyuria | Nominal | {Yes, No} |
| 6 | Water Consumption | Nominal | {Yes, No} |
| 7 | Excessive Thirst | Nominal | {Yes, No} |
| 8 | Blood Pressure (mmHg) | Numeric | {50 to 200} |
| 9 | Hyper Tension | Nominal | {Yes, No} |
| 10 | Tiredness | Nominal | {Yes, No} |
| 11 | Problem in Vision | Nominal | {Yes, No} |
| 12 | Kidney Problem | Nominal | {Yes, No} |
| 13 | Hearing Loss | Nominal | {Yes, No} |
| 14 | Itchy Skin | Nominal | {Yes, No} |
| 15 | Genetic | Nominal | {Yes, No} |
| 16 | Diabetic | Nominal | {Yes, No} |

*B. Data Preprocessing*

To achieve the goal of this research, some data preprocesses have been done on the diabetes dataset. For instance, the exact numeric value of the attributes is not meaningful to predict diabetes. As such we convert the numeric attribute values into nominal. For example, the patient's age is classified into three categories, such as Young (10-25 years), Adult (26- 50 years) and Old (above 50 years). Similarly, patient's weight is classified into three categories, such as Underweight (less than equal 40 Kgs), normal (41-60 Kgs) and Overweight (above 60 Kgs). Finally, blood pressure is classified into three categories, such as Normal (120/80 mmHg), Low (less than 80 mmHg) and High (greater than 120 mmHg).

*C. Apply Machine Learning Techmiques*

Once the data has been ready for modeling, we employ four popular *machine learning* classification techniques to predict diabetes mellitus. Hence we give an overview of these techniques.

*1) Support Vector Machines:* This is one of the most popular classification technique proposed by J. Platt et. al. [1]. A Support Vector Machine (SVM) is a excludent classifier, formally characterized the data by separating a hyperplane. SVM isolates entities in specified classes. It can also identify and classify instances which is not supported by data. SVM is not caring in the distribution of acquaring data of each class. The one extension of this algorithm is to execute regression analysis to produce a linear function and another extension is learning to rank elements to produce classification for individual elements.

*2) Naive Bayes:* Naive Bayes is a popular probabilistics classification technique proposed by John et. al. [2]. Naive Bayes also called Bayesian theorem is a simple, effective and commonly used machine learning classifier. The algorithm calculates probabilistic results by counting the frequency and combines the value given in data set. By using Bayesian theorem, it assumes that all attributes are independent and based on variable values of classess. In real world application, the conditional independence assumption rarely holds true and gives well and more sophisticate classifier results.

*3) K-Nearest Neighbor Algorithm:* K-nearest neighbor is simple classification and regression algorithm that used non parametric method proposed by Aha *et. al.* [3]. The algorithm records all valid attributes and classifies new attributes based on their resemblance measure. To determine the distance from point of interest to points in training data set it uses tree like data structure. The attribute is classified by its neighbors. In a classification technique, the value of k is always a positive integer of nearest neighbor. The nearest neighbors are chosen from a set of class or object property value.

*4) Decision Tree:* A decision tree is a tree that provides powerful classification techniques to predict diabetes mellitus. The majority of the information highlights limited discrete areas and feature called the "classification". Every discrete area and feature of the domain is called a class. An input feature of the class attribute is labeled with the internal node in a decision tree. The leaf node of the tree is labeled by attribute and each attribute associated with a target value. The highest information gain for all the attribute is calculated in each node of the tree.

There are some popular decision tree algorithms are available to classify diabetic data in machine learning techniques, including ID3, J48, C4.5, C5, CHAID and CART. In our research, C4.5 decision tree algorithm has been chosen to measure performance analysis of the diabetic data. C4.5 provides extended features of ID3 decision tree algorithm proposed by Ross Quinlan *et. al.* [4]. C4.5 decision tree uses same training data as ID3, in which learned function is introduced. The learning method can be used to diagnose medical data to predict the value of the decision attribute. In each branch node of the tree, C4.5 selects the attribute value of the data that most effectively separate the tested data into subset data which enriched the class. The tree is generated by the normalized information gain. The normalized information gain is picked to make the decision from the highest value attribute is evaluated from C4.5 decision tree.

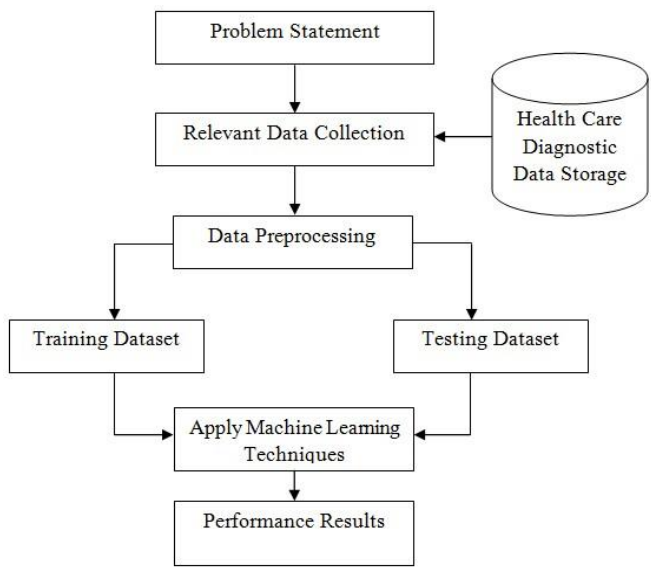

Figure 1: An overview of the overall process

Figure 1 shows an overview of the overall process of our work. According to Figure 1, after defining the problem we collect the relevant data from the Diagnostic Data Storage. We then preprocess the data for the purpose of building the prediction model. After that we apply various machine learning techniques discussed above on the training dataset. Finally, test dataset is used to measure the performance of the techniques in order to choose the best classifier for predicting diabetes mellitus.

IV. EXPERIMENTAL RESULTS AND DISCUSSION

In order to measure the performance of the classification techniques, we perform the most popular N-fold (N=10) cross validation technique. In N-fold cross validation, the dataset is separated into N-folds, where each fold is used as testing data to evaluate the model performance. The testing and training data repeats N times until completing the whole procedure. According to N-folds cross validation, we divide the data into 10-folds where each fold is nearly same with other folds in the dataset. Execution of each iteration of trained learning scheme is held out from nine folds and the performance of the learning techniques is computed using the remaining 1 fold known as testing set. Learning scheme techniques performed 10 times on training data sets and lastly the prediction accuracies are averages for 10 data sets to yield an overall prediction results in terms of precision, recall, f-measure and accuracy, which are defined below.

*A. Evaluation Metric*

In this section, we define all the metrics used in our experiments. If TP belongs to true positive rate and FP belongs to false positive rate then according to [5] the formal definition of precision is,

$$\text{Precision} = \frac{TP}{TP + FP}$$

Similarly recall is defined as below where FN represents the false negative rate [5].

$$\text{Recall} = \frac{TP}{TP + FN}$$

The F-measure can be calculated using the value of precision and recall, and defined as below [5] -

$$\text{F-measure} = \frac{2 * \text{Recall} * \text{Precision}}{\text{Precision} + \text{Recall}}$$

In addition to these metrics we also calculate the accuracy based on the correctly classify instances performed the machine learning techniques. Formally, Accuracy is calculated as below [5] -

$$\text{Accuracy} = \frac{TP + TN}{TP + TN + FP + FN}$$

*B. Comparison Results*

In order to evaluate the performance of different machine learning techniques, we have shown the prediction results in Figure 2 on the basis of precision, recall and f-measure. The figure shows the results of various machine learning techniques such as SVM, NB, KNN and C4.5. If we observe Figure 2, we see that classifier C4.5 achieves better results than other classifiers to predict diabetes mellitus. According to Figure 2, C4.5 achieves 72% precision, 74% recall and 72% f-measure on this dataset, which is greater than other learning techniques. This experimental result provides an evidence that decision tree performs well on medical datasets for the purpose of predicting diabetes mellitus based on various risk factors, dicussed in the earlier section.

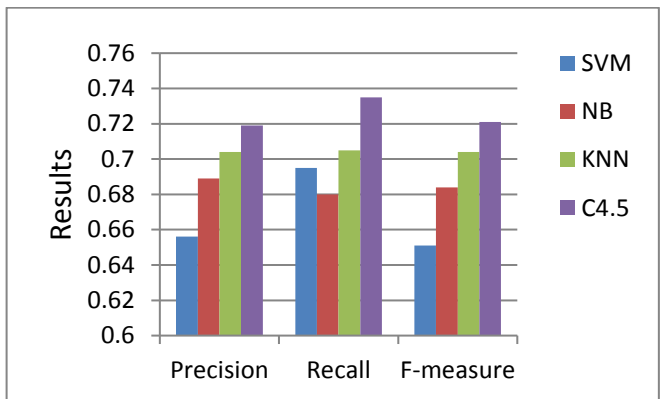

**Figure 2: Predictions Results of Various Machine Learning Techniques**

In addition to precision, recall and f-measure, we also calculate the direct accuracy rate in percentage of all these classifiers shown in Figure 3. If we observe Figure 3, we also see that C4.5 decision tree technique outperforms than other techniques to predict diabetes mellitus.

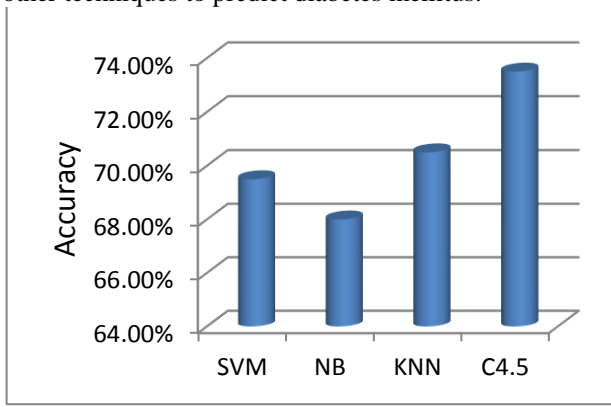

**Figure 3: Accuracy Results of Various Machine Learning Techniques**

Overall, we have chosen the best machine learning technique to predict diabetes mellitus to achieve high performance, based on the evaluation criteria discuss above. All the techniques mentioned over are estimated on an unseen testing diabetic dataset. The technique which accomplishes the highest performance in terms of precision, recall, f-measure and accuracy, is considered to be the best choice. Based on Figure 2 and Figure 3, it can be observed that C4.5 decision tree achieved the better accuracy of 73.5% to predict diabetes mellitus utilizing a given medical dataset.

## V. CONCLUSION

In this work, we have analyzed the early prediction of diabetes by taking into account various risk factors related to this disease using machine learning techniques. Extracting knowledge from real health care dataset can be useful to predict diabetic patients. To predict diabetes mellitus effectively, we have done our experiments using four popular machine learning algorithms, namely Support Vector Machine (SVM), Naive Bayes (NB), K-Nearest Neighbor (KNN) and C4.5 decision tree, on adult population data to predict diabetes mellitus.

Our experimental results shown that the performance of C4.5 decision tree is significantly superior to other machine learning techniques for the classification of diabetic data. The experimental results could assist health care to take early prevention and make better clinical decisions to control diabetes and thus save human life. We believe that our experimental study could be helpful to make a control plan for diabetes mellitus in future.

ACKNOWLEDGMENT

The authors were very grateful to Dr. Shahed Parvez, Director, Medical Centre Chittagong and Dr. Mohammad Maruf Faruqi, RMO, ICU, Medical Centre Chittagong, Bangladesh for providing the real diagnostic data for the purpose of this study.

REFERENCES

[1] Platt, John C. "12 fast training of support vector machines using sequential minimal optimization." Advances in kernel methods (1999): 185-208.

[2] John, George H., and Pat Langley. "Estimating continuous distributions in Bayesian classifiers." Proceedings of the Eleventh conference on Uncertainty in artificial intelligence. Morgan Kaufmann Publishers Inc., 1995.

[3] Aha, David W., Dennis Kibler, and Marc K. Albert. "Instance-based learning algorithms." Machine learning 6.1 (1991): 37-66.

[4] Ross Quinlan (1993). C4.5: Programs for Machine Learning. Morgan Kaufmann Publishers, San Mateo, CA.

[5] Witten, I. H. et al. (1999). Weka: Practical machine learning tools and techniques with Java implementations.

[6] Morteza, M., Franklyn, P., Bharat, S., Linying, D., Karim, K. and Aziz G. 2015. Evaluating the Performance of the Framingham Diabetes Risk Scoring Model in Canadian Electronic Medical Records. Canadian journal of diabetes 39, 30(April. 2015), 152-156.

[7] V., A. K. and R., C. 2013. Classification of Diabetes Disease Using Support Vector Machine. International Journal of Engineering Research and Applications. 3, (April. 2013), 1797-1801.

[8] Carlo, B G., Valeria, M. and Jesús, D. C. 2011. The impact of diabetes mellitus on healthcare costs in Italy. Expert review of pharmacoeconomics & outcomes research. 11, (Dec. 2011),709-19.

[9] Nahla B., Andrew et al. 2010. Intelligible support vector machines for diagnosis of diabetes mellitus. Information Technology in Biomedicine, IEEE Transactions. 14, (July. 2010), 1114-20.

[10] Abdullah A. Aljumah et al., Application of data mining: Diabetes health care in young and old patients, Journal of King Saud University - Computer and Information Sciences, Volume 25, Issue 2, July 2013, Pages 127-136

[11] Kavakiotis, Ioannis, Olga Tsave, AthanasiosSalifoglou, NicosMaglaveras, IoannisVlahavas, and IoannaChouvarda. "Machine learning and data mining methods in diabetes research." Computational and structural biotechnology journal (2017).

[12] Zheng, Tao et al. "A machine learning-based framework to identify type 2 diabetes through electronic health records." International journal of medical informatics 97 (2017): 120- 127.

[13] Rani, A. Swarupa, and S. Jyothi. "Performance analysis of classification algorithms under different datasets." In Computing for Sustainable Global Development (INDIACom), 2016 3rd International Conference on, pp. 1584-1589. IEEE, 2016.

[14] Kandhasamy, J. Pradeep, and S. Balamurali. "Performance analysis of classifier models to predict diabetes mellitus." Procedia Computer Science 47 (2015): 45-51.

[15] Y. Huang, P. McCullagh, N. Black, R. Harper, Feature selection and classification model construction on type 2 diabetic patients'data, Artificial Intelligence in Medicine 41 (3) (2015) 251–262.

[16] Meng, X. H., Huang, Y. X., Rao, D. P., Zhang, Q., & Liu, Q. (2013). Comparison of three data mining models for predicting diabetes or prediabetes by risk factors. The Kaohsiung journal of medical sciences, 29(2), 93-99.